\documentclass[runningheads]{llncs}

\usepackage[T1]{fontenc}
\usepackage[utf8]{inputenc}
\usepackage{float}
\usepackage{standalone}

\usepackage{graphicx}
\usepackage{amsmath}
\usepackage{amssymb}
\usepackage{hyperref}
\usepackage{booktabs}
\usepackage{float}
\usepackage{xcolor}
\usepackage{bbding} % For envelope symbol to mark corresponding author
\usepackage{tikz}

\usetikzlibrary{
  shapes,
  arrows.meta,
  positioning,
  calc,
  backgrounds,
  fit,
  decorations.pathreplacing,
  patterns
}
\usepackage{xcolor}
\usepackage{amssymb}
\usepackage{sfmath}
\renewcommand{\familydefault}{\sfdefault}

\definecolor{plausibleColor}{HTML}{81C784}   % Softer green
\definecolor{implausibleColor}{HTML}{E57373} % Softer red
\definecolor{layerColor}{RGB}{193, 205, 224} % Light gray-blue
\definecolor{decoderColor}{HTML}{FFD54F}     % Golden yellow
\definecolor{prColor}{HTML}{E1F0F5}          % Light gray-blue for better readabilit
\usepackage{color}

\begin{document}

\title{In Machina N400: Pinpointing Where a Causal Language Model Detects Semantic Violations}

\titlerunning{In Machina N400}
% Use abbreviated title if the full title is too long for running head

\author{Christos-Nikolaos Zacharopoulos\Envelope\orcidID{0000-0002-3430-5115} \and
Revekka Kyriakoglou\inst{2}\orcidID{0009-0007-8626-8401}}

% The \Envelope symbol marks the corresponding author

%\authorrunning{C. Zacharopoulos and R. Kyriakoglou}

\institute{Independent Researcher\\
\email{christonik@gmail.com}
\and
 Université Paris 8 Vincennes - Saint-Denis, Paris, France\\
\email{revekka.kyriakoglou@univ-paris8.fr}}
\maketitle

\begin{abstract}
How and where does a transformer notice that a sentence has gone semantically off the rails?
To explore this question, we evaluated the causal language model \(\phi\)-2 (phi-2) using a carefully curated corpus, with sentences that concluded plausibly or implausibly. Our analysis focused on the hidden states sampled at each model layer. To investigate how violations are encoded, we utilized two complementary probes. First, we conducted a per-layer detection using a linear probe. Our findings revealed that a simple linear decoder struggled to distinguish between plausible and implausible endings in the lowest third of the model's layers. However, its accuracy sharply increased in the middle blocks, reaching a peak just before the top layers. Second, we examined the effective dimensionality of the encoded violation. Initially, the violation widens the representational subspace, followed by a collapse after a mid-stack bottleneck. This might indicate an exploratory phase that transitions into rapid consolidation.
Taken together, these results contemplate the idea of alignment with classical psycholinguistic findings in human reading, where semantic anomalies are detected only after syntactic resolution, occurring later in the online processing sequence.

\end{abstract}

\section{Introduction}
Why does a single anomalous word, a "dog" floating in a coffee cup, immediately jolts the mind? From the moment a sentence unfolds, the cerebral cortex generates a cascade of probabilistic hypotheses, continuously comparing incoming input with an ever-updating internal model of the linguistic and conceptual world. This view, rooted in Bayesian and predictive-coding theories of perception and action \cite{Friston2005ATheory}, has been powerfully confirmed in the language domain by four decades of electrophysiology: semantic violations reliably summon a negative deflection around 400 ms, the celebrated N400, indexing the brain's rapid computation of a "prediction error" when reality collides with prior expectation \cite{Kutas1980ReadingSS,Kutas2011ThirtyYA}. 

Yet an essential question remains: \emph{what internal code carries these predictions?} With the advent of transformer-based Neural Language Models (NLMs) \cite{Vaswani2017AttentionIA}, a unique methodological opportunity has emerged. These models are the first engineered systems to generate open-ended, human-like text with sufficient fluency to let us inspect their hidden representations as a window onto putative computations in the biological brain. A growing corpus of work shows striking isomorphisms between activity patterns in deep language models and those measured with fMRI, MEG, and intracranial recordings along the human perisylvian network \cite{caucheteux2022brains,caucheteux2022deep,goldstein2022shared,Kuribayashi2023HumanLike}. In effect, NLMs constitute an \emph{in-machina} laboratory where hypotheses about cortical algorithms, predictive coding, and hierarchical abstraction can be spelled out, manipulated, and tested at a level of mechanistic detail rarely attainable \emph{in-vivo}.
At the same time, caution is warranted. Fluency alone does not imply shared computation: current NLMs lack grounded semantics, sensorimotor constraints, and developmental trajectories that shape the human language faculty. 

Prior computational works have mainly relied on coarse behavioral proxies such as next-word probability or surprisal calculations. These measures, while helpful, flatten the rich temporal dynamics revealed as the input token's vector is successively transformed into new representations across later layers. To move beyond this impasse, we adopt a neuro-computational strategy: we interrogate the internal activity of a modern transformer (Phi-2; \cite{javaheripi2023phi}) with stimuli precisely designed to illuminate the N400 in humans.
Specifically, we create sentence pairs of equal length that terminate in either a semantically plausible or an impossible final word (e.g., "He drinks coffee with cream and \emph{dog}").\footnote{For a selection of stimuli used in the experiment, please see Appendix \ref{appendixA}} This paradigm allows us to track, layer by layer, how prediction error propagates through the model's attention heads and feed-forward sublayers.

To that end, we capture the hidden state at each successive layer, normalize the ensuing vectors, and ask a simple question: can a linear decoder read out "anomaly or not" above chance? We then chart this signal's emergence layer by layer and validate its robustness with cluster-based permutation statistics. A complementary analysis of participation ratio allows us to disentangle dimensionality expansion from pure prediction error. In short, we treat the transformer as a transparent, time-unfolding cortex and probe its state at every computational slice.

\begin{figure}[h]
    \centering
    \begin{tikzpicture}[
    >=Latex,
    line width=0.6pt,
    node distance=0.6cm and 2.5cm,
    sentence/.style={
        rectangle,
        rounded corners=6pt,
        draw,
        text width=13em,
        minimum height=2.8em,
        inner xsep=1em,
        inner ysep=0.8ex,
        font=\small,
        align=center
    },
    layer/.style={
        rectangle,
        draw,
        fill=layerColor,
        text width=9em,
        text centered,
        minimum height=2.2em,
        font=\small
    },
    probe/.style={
        rectangle,
        draw,
        rounded corners=4pt,
        text width=10em,
        align=center,
        minimum height=5.5em,
        font=\small,
        inner sep=0.6em
    },
    fwd/.style={draw,-{Latex},line width=0.6pt},
    probe_lr/.style={draw,densely dashed,line cap=round,line width=1.5pt, color=decoderColor!80!black},
    probe_pr/.style={draw,loosely dashed,line cap=round,line width=1.5pt, color=prColor!50!black},
    panellabel/.style={
        font=\Large\bfseries,
        inner sep=0pt
    }
]

% Panel labels at top - aligned horizontally
\node[panellabel] at (0, 4) {(A)};
\node[panellabel] at (7, 4) {(B)};
\node[panellabel] at (14.5, 4) {(C)};

% ==================================================================
% SECTION A: INPUT
% ==================================================================
\node[sentence, fill=plausibleColor] (plausible) at (0, 1.5)
  {``The baker near the school bakes \textbf{bread}.''};
\node[below=0.05cm of plausible, font=\small] {Plausible};
\node[sentence, fill=implausibleColor, below=0.8cm of plausible] (implausible)
  {``The baker near the school bakes \textbf{sunlight}.''};
\node[below=0.05cm of implausible, font=\small] {Implausible};
\node[above=0.4cm of plausible, text width=13em, font=\small, align=center]
  {Matched pairs differing\\in final noun};

% ==================================================================
% SECTION B: TRANSFORMER
% ==================================================================
\node[layer, right=2.8cm of plausible, yshift=0.4cm] (layer1)
  {Layer 1};
\node[layer, below=0.4cm of layer1] (layer2)
  {Layer 2};
\node[layer, below=0.4cm of layer2] (layer3)
  {Layer 3};
\node[below=0.3cm of layer3, font=\large] (dots) {$\vdots$};
\node[layer, below=0.3cm of dots] (layer32)
  {Layer 32};

% Transformer box
\begin{pgfonlayer}{background}
\node[
  draw,
  rounded corners=6pt,
  inner xsep=0.9cm,
  inner ysep=0.7cm,
  line width=0.7pt,
  fit=(layer1) (layer32)
] (transformerBox) {};
\end{pgfonlayer}
\node[font=\small\bfseries, above=0.15cm of transformerBox.north]
  {Phi-2 Transformer};
\node[font=\scriptsize, below=0.05cm of transformerBox.north]
  {(32 layers)};

% Forward arrows
\foreach \L in {layer1,layer2,layer3,layer32}{
  \path[fwd] (plausible.east) -- ++(0.5,0) |- (\L.west);
  \path[fwd] (implausible.east) -- ++(0.5,0) |- (\L.west);
}

% ==================================================================
% SECTION C: ANALYSIS WITH VISUALS
% ==================================================================
% Classifiability probe with logistic regression visualization
\node[probe, fill=decoderColor, right=2.8cm of layer1, yshift=-0.5cm] (decoder)
  {\textbf{Classifiability}\\[0.3ex]
   \begin{tikzpicture}[scale=0.5, baseline=(current bounding box.center)]
     % Plausible points (green)
     \foreach \x/\y in {0.5/1.2, 0.8/1.5, 0.3/1.8, 0.6/2.1, 0.9/1.0}{
       \fill[plausibleColor!80!black] (\x,\y) circle (5pt);
     }
     % Implausible points (red)
     \foreach \x/\y in {2.0/0.3, 2.3/0.6, 1.7/0.5, 2.2/0.2, 1.9/0.8}{
       \fill[implausibleColor!80!black] (\x,\y) circle (5pt);
     }
     % Separating hyperplane
     \draw[thick, blue] (0.2,0) -- (2.8,2.5);
     % Sigmoid curve for logistic regression
     \draw[thick, red] plot[domain=0.2:2.8, samples=100] (\x, {2.5 / (1 + exp(-5*(\x-1.5)))});
     \node[font=\tiny\bfseries, color=red] at (2.0, 2.0) {Logistic Regression};
   \end{tikzpicture}\\[-0.2ex]
   {\scriptsize Logistic regression separation}};
\node[right=0.2cm of decoder.east, font=\small] {$\rightarrow$ AUC$_\ell$};

% Dimensionality probe with manifold visualization and PR formula
\node[probe, fill=prColor, below=1.8cm of decoder] (pr)
  {\textbf{Dimensionality}\\[0.3ex]
   \begin{tikzpicture}[scale=0.5, baseline=(current bounding box.center)]
     % High-dimensional representation (green)
     \draw[thick, fill=plausibleColor!40!white, draw=plausibleColor!80!black]
       plot[smooth cycle, tension=0.8] coordinates {
         (-0.5,1.5) (0.5,2.5) (1.5,2.0) (2.0,1.0) (1.0,0.0) (0.0,0.5)
       };
     \node[font=\small\bfseries, color=plausibleColor!80!black] at (0.5, 1.5) {Plaus.};
     % Lower-dimensional representation (red)
     \draw[thick, fill=implausibleColor!40!white, draw=implausibleColor!80!black]
       (2.5, 1.0) ellipse (0.5 and 0.8);
     \node[font=\small\bfseries, color=implausibleColor!80!black] at (2.5, 1.8) {Implaus.};
   \end{tikzpicture}\\[-0.2ex]
   {\scriptsize $\text{PR} = \frac{\left(\sum_i \lambda_i\right)^2}{\sum_i \lambda_i^2}$}};
\node[right=0.2cm of pr.east, font=\small] {$\rightarrow$ $\Delta$PR$_\ell$};

% Probe connections (thicker and colored to match boxes)
\coordinate (decoderAnchor) at ([xshift=-0.1cm]decoder.west);
\coordinate (prAnchor) at ([xshift=-0.1cm]pr.west);
\foreach \L in {layer1,layer2,layer3,layer32}{
  \path[probe_lr] (\L.east) to[out=0, in=180] (decoderAnchor);
  \path[probe_pr] (\L.east) to[out=0, in=180] (prAnchor);
}

% ==================================================================
% LEGEND (bottom, centered)
% ==================================================================
\node[font=\small, below=1.3cm of transformerBox.south] (legend_note)
  {\textit{Metrics computed at each layer} $\ell = 1, \ldots, 32$};
\coordinate (leg_center) at ([yshift=-0.4cm]legend_note.south);
\draw[fwd] ([xshift=-2.5cm]leg_center) -- ++(0.8,0)
  node[right, font=\scriptsize] {forward};
\draw[probe_lr] ([xshift=-0.2cm]leg_center) -- ++(0.8,0)
  node[right, font=\scriptsize] {classif. probe};
\draw[probe_pr] ([xshift=2.5cm]leg_center)  -- ++(0.8,0)
  node[right, font=\scriptsize] {dimen. probe};

\end{tikzpicture}
\caption{
\textbf{Overview of the analysis pipeline.}
\textbf{(A)} Stimuli consist of 1520 matched sentence pairs (760 plausible, 760
implausible), each differing only in the final noun while preserving a fixed
syntactic template.  
\textbf{(B)} Each sentence is processed once through all 32 layers of the
Phi-2 transformer. For every layer $l$, the hidden-state matrix
$H^{(l)}(x) \in \mathbb{R}^{T_x \times d}$ is extracted and reshaped into a
vector $h^{(l)}(x) = \operatorname{vec}(H^{(l)}(x))$. Activations are normalized
within each layer across sentences prior to feature computation.  
\textbf{(C)} Two complementary layer-wise metrics are computed.
(i) A logistic-regression decoder trained on the mean activation
$\mu^{(l)}(x)$ yields a decoding score $\mathrm{AUC}_l$, quantifying how well
layer $l$ separates plausible from implausible sentences.  
(ii) The representational dimensionality of each layer is estimated using the
participation ratio $\mathrm{PR}_l$, derived from the eigenvalues of the
covariance matrix of $h^{(l)}(x)$ across sentences.  
Together, these analyses characterize how semantic-violation information becomes
linearly decodable and how the effective dimensionality of the model’s internal
representations evolves across layers.
}
 \label{fig:conceptual}
\end{figure}
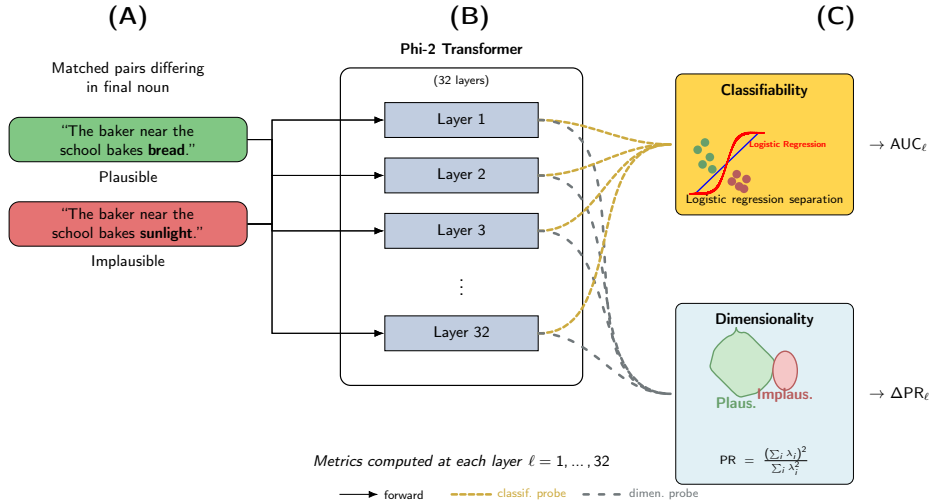

\emph{This work addresses a precise question: At which computational depth does a transformer first register a semantic anomaly, and does this temporal pattern align with the timing of the human N400 response?} If semantic anomaly detection emerges only in mid-to-late layers of the network, this would parallel influential models of human sentence processing in which semantic integration occurs after initial structural analysis \cite{friederici2002towards,Kuperberg2007}. Such a correspondence would suggest that both artificial and biological language systems postpone full semantic evaluation until sufficient syntactic constraints have been established. We investigate whether a homologous time course unfolds, not across milliseconds, but across the successive algorithmic steps of a language model. Figure~\ref{fig:conceptual} provides a conceptual overview of our approach.

\section{Methods}

\paragraph{Activation Extraction and Stimuli.}
Sentences from a rigorously controlled stimulus set were processed using the
transformer-based causal language model \textit{Phi-2} (2.7B parameters),
trained for next-token prediction \cite{javaheripi2023phi}. a causal (autoregressive) architecture because its attention mask enforces a left-to-right information flow, ensuring that each token is predicted based only on preceding context, enabling a comparison with the temporal dynamics of human reading that
unfolds over time. The model size was chosen to balance computational
tractability with sufficient representational capacity for semantic processing,
while remaining accessible for detailed layer-wise analysis. Phi-2 was used in
evaluation mode with deterministic forward passes (no gradient computation, 
temperature=1.0, no sampling) to ensure reproducible activations across all 
analyses.

Each sentence followed the fixed template
\[
\text{``The [subject] [location] [verb] [object].''}
\]
such that the grammatical structure, length, and word order were identical
across the entire stimulus set. All sentences were matched at the word level 
(identical number of words per sentence), ensuring structural uniformity. 
The semantic anomaly was always positioned in the final word. Subword 
tokenization introduced minor variation in token counts (typically ±1--2 tokens 
per sentence), but this variation was negligible and did not introduce 
confounding effects related to sequence truncation or padding, as all sentences 
remained well within the model's context window.

The stimulus set was generated from a structured lexicon that mapped human
professions to verbs, expected objects, and corresponding violation objects. For
example, the lexicon defined ``baker'' with the verb ``bakes,'' an expected
object ``bread,'' and a violation object ``sunlight,'' permitting matched pairs
such as \emph{"The baker near the school bakes bread."} versus \emph{"The baker
near the school bakes sunlight."}. The full dataset comprised 1520 sentences,
exactly half containing a semantic violation.

\paragraph{Feature Extraction and Selection.}
Each sentence $x$ was tokenised with $\phi_2$'s native tokenizer and passed once
through the network. For each transformer layer $l \in \{1,\dots,32\}$, the
model produced a hidden-state matrix
\[
H^{(l)}(x) \in \mathbb{R}^{T_x \times d},
\]
where $T_x$ is the (tokenised) sequence length and $d$ the model dimension. We
reshaped $H^{(l)}(x)$ into a single vector,
\[
h^{(l)}(x) = \operatorname{vec}\!\left(H^{(l)}(x)\right) \in \mathbb{R}^{T_x d},
\]
via row-major concatenation of token embeddings.

From $h^{(l)}(x)$ we computed five distributional moments—the mean, median,
variance, skewness, and kurtosis—yielding a 5-dimensional descriptor
\[
m^{(l)}(x)
  = \bigl[\mu^{(l)}(x), \operatorname{med}^{(l)}(x), \operatorname{var}^{(l)}(x),
     \operatorname{skew}^{(l)}(x), \operatorname{kurt}^{(l)}(x)\bigr].
\]

To determine whether higher-order moments contributed additional predictive
signal beyond the mean, we trained $\ell_2$-regularised logistic-regression
decoders on (i) the mean alone and (ii) every non-empty subset of the five
statistics, using stratified $k$-fold cross-validation. No subset achieved
statistically reliable improvements over the mean-only probe. Supplementing the
mean with higher-order moments neither improved generalisation nor stabilised
the probe weights, and substantially increased the spectral condition number of
the design matrix. These results established the mean activation as both
\emph{sufficient} and \emph{stable} for decoding, motivating its exclusive use
in downstream analyses.

\paragraph{Normalization and Data Aggregation.}
Before feature computation, each layer's activations were normalised separately
across sentences. For every layer $l$, we applied robust centering and scaling:
\[
\tilde{h}^{(l)}(x)
  = \frac{h^{(l)}(x) - \operatorname{median}(h^{(l)})}
         {\operatorname{IQR}(h^{(l)})},
\]
where the median and interquartile range (IQR) were computed across the entire
sentence set. This procedure mitigated the influence of lexical outliers and
reduced sensitivity to morphological variability.

\paragraph{Layer-wise Decoding Analysis.}
Following standard neural decoding methodology \cite{king2014characterizing}, we
trained an $\ell_2$-regularised logistic regression classifier for each layer
$l$, using only the mean activation $\mu^{(l)}(x)$ as the predictive feature.
For each layer, the classifier learned parameters $(w^{(l)}, b^{(l)})$ by
minimising
\[
\mathcal{L}(w,b)
  = -\sum_x \Bigl[
      y(x)\log \sigma(w\mu^{(l)}(x) + b)
      + (1 - y(x))\log\!\bigl(1 - \sigma(w\mu^{(l)}(x) + b)\bigr)
    \Bigr]
    + \lambda \|w\|_2^2,
\]
with $\lambda = 1$. Performance was evaluated using the ROC-AUC averaged across
the five cross-validation folds, yielding a layer-specific decoding score
$\mathrm{AUC}_l$.

\paragraph{Cluster-Based Permutation Testing.}
To assess whether decoding performance at each layer exceeded chance while
controlling the family-wise error rate, we applied a one-dimensional
cluster-based permutation test following \cite{maris2007nonparametric}. For each
layer $l$, we computed a one-sample $t$-statistic comparing fold-level
decoding scores to the chance level of 0.5,
\[
t_l = t(\mathrm{AUC}_l - 0.5),
\]
with $df = 4$. Layers with $|t_l| > 2.78$ (two-tailed $p<0.05$) formed
supra-threshold clusters along the layer dimension. The cluster statistic was
the sum of $t_l$ within each cluster.

A null distribution was generated using 1000 sign-flip permutations across
layers. For each permutation, we recomputed the cluster statistics and recorded
the maximum absolute cluster value. Observed clusters were assigned corrected
$p$-values equal to the proportion of permutation maxima exceeding their
statistic. Clusters with corrected $p<0.01$ were deemed significant.

\paragraph{Participation Ratio Computation.}
To quantify the effective dimensionality of activations at each layer, we
computed the participation ratio (PR), a measure commonly used in theoretical
and systems neuroscience \cite{gao2017theory,desbordes2023dimensionality}. For
each layer $l$, we formed the covariance matrix
\[
\Sigma^{(l)} = \operatorname{Cov}\!\left(h^{(l)}(x)\right)
\]
across sentences. Let $\lambda^{(l)}_1,\dots,\lambda^{(l)}_n$ denote its
eigenvalues. The participation ratio is
\[
\mathrm{PR}_l
  = \frac{\bigl(\sum_{i=1}^{n} \lambda^{(l)}_i\bigr)^2}
         {\sum_{i=1}^{n} (\lambda^{(l)}_i)^2},
\]
providing a scalar estimate of the layer's representational dimensionality.
\section{Results}

\subsection{Decoding accuracy across layers}
\begin{figure}
    \centering
    \includegraphics[width=0.8\textwidth]{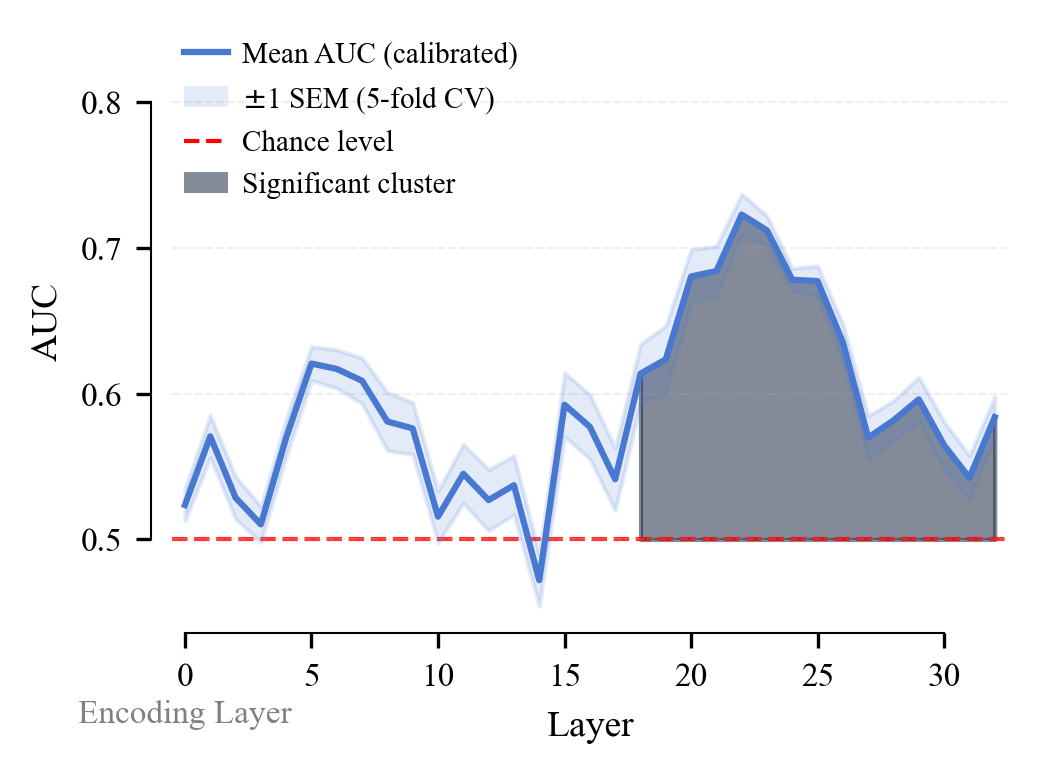}
\caption{%
\textbf{Layer-wise decoding of semantic anomalies in \textit{Phi-2}.}  
Mean ROC–AUC (blue; ±1 SEM shaded) for a logistic classifier that distinguishes plausible from violation endings at each encoding layer.  
The red dashed line shows chance performance (0.5).  
The grey shading marks the only cluster of consecutive layers (18–30) whose AUC reliably exceeded chance after cluster-based permutation correction ($p<0.001$).}
\label{fig:auc_layers}
\end{figure}

\begin{figure}
    \centering
    \includegraphics[width=0.8\textwidth]{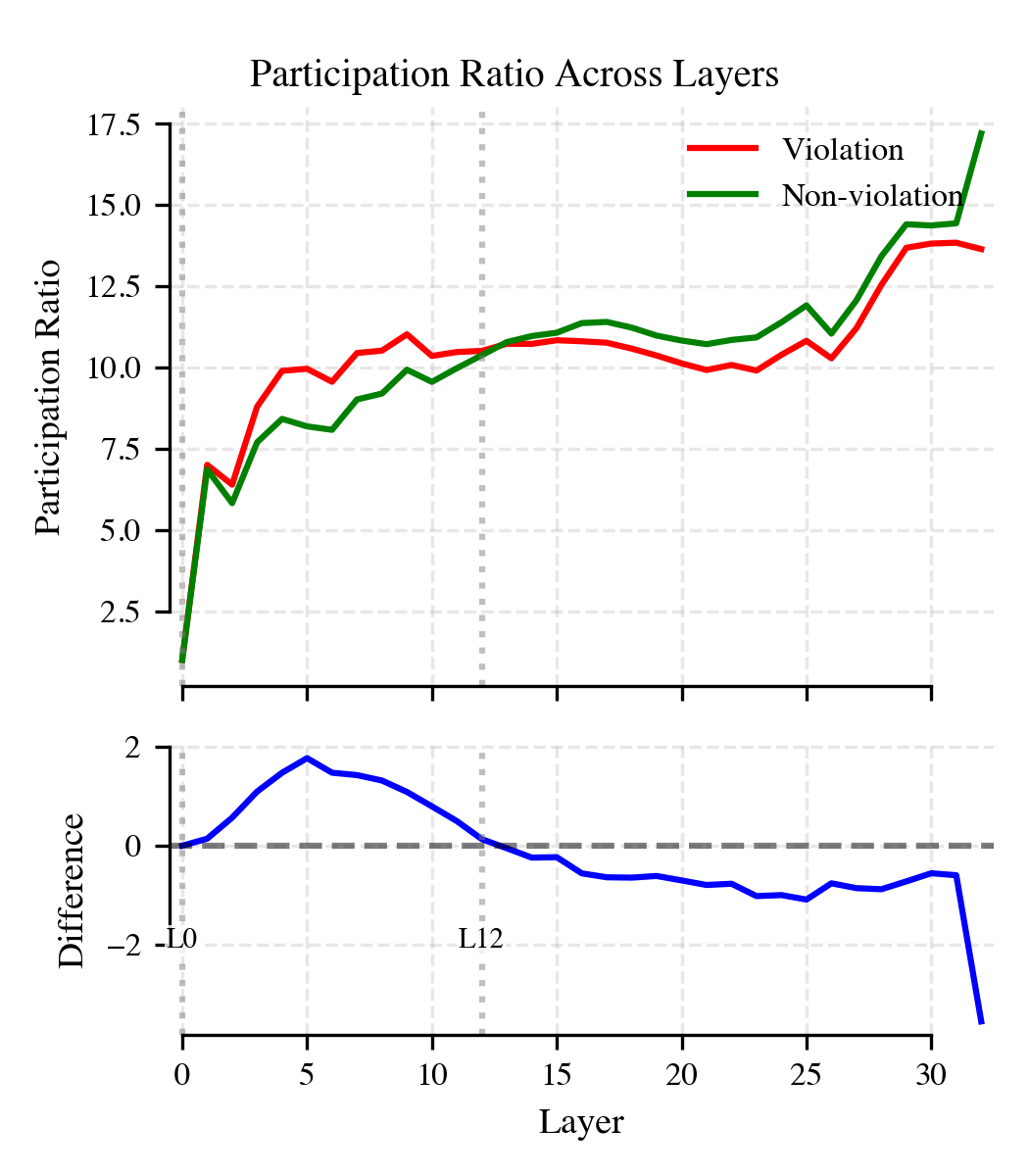}
\caption{%
\textbf{Effective dimensionality of hidden states.}  
\emph{Top:} Participation ratio (PR) for violation (red) and control (green) sentences across the 32 encoding layers of \textit{Phi-2}.  \emph{Bottom:} Difference trace (violation – control) relative to the zero baseline (grey dashed line).  Violations initially occupy a higher‐dimensional subspace (layers 1–6), converge with controls around the mid-stack bottleneck (layer 12, grey vertical tick), and become marginally more compressed in deeper layers.  These dynamics suggest an early expansion of representational space to accommodate unexpected input, followed by a gradual re-integration as contextual constraints accumulate.}
\label{fig:pr_layers}
\end{figure}

Figure~\ref{fig:auc_layers} shows the mean ROC–AUC obtained by layer-wise logistic classifiers.  
A cluster-based permutation test identified one contiguous cluster (layers 18–30) whose AUC values exceeded the chance level of 0.5 (\emph{cluster} $p<0.01$).  
Within this cluster, peak decoding occurred at layer 22 (mean $AUC=0.723$, $SEM=0.014$).

\subsection{Participation ratio of hidden activations}

Participation ratios (PR) for violation and control conditions are plotted in Figure~\ref{fig:pr_layers}.  
Across the lower stack (layers 1–6), the violation PR exceeded the control PR (maximum difference $=1.9$ dimensions at layer 5). The two curves converged at layer 12.  
The mean PR difference after the inflection point remained negative.

\section{Discussion}

\subsection{A late, layer‐bound marker of semantic error}

Our layer-wise decoding shows that \textit{Phi-2} registers a semantic anomaly with discriminability peaking in a narrow band around layers 22--24. Translating depth into real time is necessarily indirect, yet a negative result is already informative: \emph{the signal is absent from the first layers}. In other words, semantic error detection emerges only after the network has traversed a substantial portion of its hierarchy, rather than in the shallow stages that would more closely resemble early EEG components such as the ELAN\footnote{Early Left Anterior Negativity (ELAN): an ERP component in language processing, typically occurring 100--300 ms after stimulus onset, associated with early syntactic processing \cite{friederici2002towards}.} or LAN.\footnote{Left Anterior Negativity (LAN): an ERP component with a slightly later time course (300--500 ms), also associated with syntactic processing \cite{molinaro2011grammatical}.} \cite{Kutas1980ReadingSS,Kutas2011ThirtyYA,KuperbergReview2020}. This pattern points to the view that both artificial and biological language systems postpone semantic anomaly detection until a sufficiently broad context has been integrated.

%\subsection{Two phases of representational geometry}
%The participation‐ratio analysis uncovers a biphasic trajectory. In the lower stack, violations expand the effective dimensionality of hidden states, suggesting that the network transiently recruits additional representational axes to search for a compatible interpretation. The mid‐stack convergence at layer $\sim$12 marks a bottleneck after which dimensionality for the anomalous sentences becomes  \emph{smaller} than for canonical ones. We interpret this compression as the network settling into a lower‐entropy state once the improbability of the input has been resolved, and the result is a stable prediction error. Such a transition from exploratory expansion to confirmatory compression is reminiscent of a reanalysis similar to this observed in human data \cite{Kuperberg2007,Brouwer2017,BornkesselSchlesewsky2008,Kaan2000,Osterhout1994}.

\subsection{Two phases of representational geometry}
Participation–ratio (PR) analysis revealed a biphasic trajectory. In the lower stack, semantic violations \emph{increased} the effective dimensionality of hidden states compared with well-formed controls, indicating a transient dispersion of activity across a broader subspace. Around layer~$\sim12$, this trend reversed: beyond that point, the PR for anomalous sentences fell below that of canonical sentences, producing a dimensionality cross-over. We interpret the initial expansion as the network distributing activation across additional directions while integrating conflicting cues, and the subsequent contraction as the computation coalescing onto a smaller set of dominant directions once a stable—albeit low-probability—interpretation is reached. Crucially, the late-stage drop in PR does not imply information loss; it instead signals increased \emph{anisotropy}, with variance becoming concentrated along fewer principal axes.  

This expansion–contraction profile aligns qualitatively with reanalysis signatures in human neurophysiology, where an early, diffuse response to anomalies is followed by a more focal, sustained pattern (e.g.\ the N400–P600 sequence) \cite{Kuperberg2007,Brouwer2017,BornkesselSchlesewsky2008,Kaan2000,Osterhout1994}. While the mapping from PR to ERP components is indirect, the shared time-course supports the view that both systems first explore a wider representational space before converging on a compact error state.

\paragraph{Cautious brain--model comparisons and their promise.}
The overlap between \textit{Phi-2}'s late anomaly signal and the human N400 is suggestive, but shared \emph{latency} does not guarantee shared \emph{mechanisms}. Human comprehension recruits multimodal context, pragmatic inference, and recurrent circuitry that present-day transformers largely lack. Robust tests must therefore pair layer activations with time-resolved neural data (EEG/MEG/ECoG) on identical stimuli and compare them through representational‐similarity or related analyses. Such work can expose which semantic computations truly inhabit a common latent space and which gaps reveal biological inductive biases that could guide richer model design. In turn, a principled bridge between brains and machines would give cognitive neuroscience explicit algorithms while steering machine learning toward systems that reason and, one day, may understand more like us.

\subsection{Limitations}

\paragraph{Limitations and future directions.}
The present analysis is deliberately focused and, as a consequence, leaves several open questions that define a clear research agenda.  
\emph{Model scope.}  All results were obtained with a single 2.7-B-parameter causal transformer (\textit{Phi-2}).  Because depth, scale, and training objective influence internal dynamics, the late-emerging anomaly signal and the biphasic expansion–compression pattern should be replicated in both larger autoregressive LMs and bidirectional encoders before being treated as architecture-general principles.  Future work should systematically explore how these patterns vary across models with different depths (both deeper networks with more layers and shallower networks with fewer layers) to determine whether the mid-to-late layer emergence of semantic anomaly detection is a general property of transformer architectures or specific to certain depth ranges.
\emph{Stimulus design.}  We used tightly matched sentences in which the violation always appeared on the final token.  This maximises control but also creates a positional regularity that the network could exploit.  Future work should randomise violation position, introduce graded expectancy manipulations, and explore multi-word incongruities to verify position- and context-invariance.  
 
\emph{Geometry metrics.}  We characterised representational geometry via participation ratio only; complementary measures (e.g.\ intrinsic dimensionality, manifold capacity, centre-of-mass drift) may uncover additional structure, especially during the early expansion phase.  
\emph{Statistical robustness.}  Significance was assessed with five-fold cross-validation and cluster-based permutation tests; larger held-out corpora and replication on an external stimulus set would guard against fold-wise lexical leakage and confirm generality.  
\emph{Probe methodology.} We employed linear logistic regression 
as our primary decoding method; alternative approaches such as 
non-linear classifiers, clustering analysis, or representational 
similarity analysis could reveal additional structure in how 
semantic violations are encoded across layers.

\subsubsection*{Acknowledgments.} The author acknowledges using computational resources and open-source software libraries that made this research possible.

\subsubsection*{Disclosure of Interests.} The author has no competing interests to declare relevant to this article's content.

\clearpage
\appendix\label{appendixA}
\section{Examples of Stimuli}
\begin{table}[htbp]
\centering
\begin{tabular}{ll}
\toprule
Normal Sentences & Violation Sentences \\
\midrule
The painters at the market create murals. & The painters at the market create gravity. \\
The farmer near the school harvests vegetables. & The farmer near the school harvests moonlight. \\
The builders by the river construct structures. & The builders by the river construct whispers. \\
The author near the park writes books. & The author near the park writes electricity. \\
The vets near the park treat animals. & The vets near the park treat buildings. \\
The actors near the school sing arias. & The actors near the school sing molecules. \\
The chef at the market cooks meals. & The chef at the market cooks clouds. \\
The doctors at the market cure elders. & The doctors at the market cure unicorns. \\
The builders at the hospital construct structures. & The builders at the hospital construct whispers. \\
The lawyer near the park represents clients. & The lawyer near the park represents mountains. \\
\bottomrule
\end{tabular}
\caption{Example sentences}
\label{tab:sentences}
\end{table}

% Bibliography style for LLNCS
\bibliographystyle{splncs04}
\bibliography{ref}

\end{document}